\documentclass[10pt,twocolumn,letterpaper]{article}
\usepackage{iccv}
\usepackage[accsupp]{axessibility}
\usepackage{times}
\usepackage{epsfig}
\usepackage{graphicx}
\usepackage{amsmath}
\usepackage{amssymb}
\usepackage{multirow}
\usepackage{color}
\usepackage{colortbl}
\usepackage[table]{xcolor}
\definecolor{maroon}{cmyk}{0,0.87,0.68,0.32}
\definecolor{myblue}{rgb}{0.8,0.95,0.95}
\usepackage[pagebackref=true,breaklinks=true,letterpaper=true,colorlinks,bookmarks=false]{hyperref}
\newcommand{\PreserveBackslash}[1]{\let\temp=\\#1\let\\=\temp}
\newcolumntype{C}[1]{>{\PreserveBackslash\centering}p{#1}}
\usepackage{graphicx}

\iccvfinalcopy

\def\iccvPaperID{2397} 

\ificcvfinal\fi

\begin{document}
\title{The Devil is in the Task: Exploiting Reciprocal\\Appearance-Localization Features for Monocular 3D Object Detection}

\author{Zhikang Zou$^1$\thanks{indicates equal contribution} 
        \quad Xiaoqing Ye$^1$\footnotemark[1] 
        \quad Liang Du$^2$\footnotemark[1] 
        \quad Xianhui Cheng$^4$\footnotemark[1] 
        \quad Xiao Tan$^1$ \\
        \quad Li Zhang$^3$
        \quad Jianfeng Feng$^2$
        \quad Xiangyang Xue$^4$ 
        \quad Errui Ding$^1$\\[1em]
        $^1$ Baidu Inc., China \\
        $^2$ Institute of Science and Technology for Brain-Inspired Intelligence, Fudan University,\\
        MOE Key Laboratory of Computational Neuroscience and Brain-Inspired Intelligence, Fudan University\\
        $^3$ School of Data Science, Fudan University\\
        $^4$ School of Computer Science, Fudan University\\
        }

\maketitle
\ificcvfinal\thispagestyle{empty}\fi

\begin{abstract}
Low-cost monocular 3D object detection plays a fundamental role in autonomous driving, whereas its accuracy is still far from satisfactory. In this paper, we dig into the 3D object detection task and reformulate it as the sub-tasks of object localization and appearance perception, which benefits to a deep excavation of reciprocal information underlying the entire task. We introduce a Dynamic Feature Reflecting Network, named DFR-Net, which contains two novel standalone modules: (i) the Appearance-Localization Feature Reflecting module (ALFR) that first separates task-specific features and then self-mutually reflects the reciprocal features; (ii) the Dynamic Intra-Trading module (DIT) that adaptively realigns the training processes of various sub-tasks via a self-learning manner. Extensive experiments on the challenging KITTI dataset demonstrate the effectiveness and generalization of DFR-Net. We rank 1$^{st}$ among all the monocular 3D object detectors in the KITTI test set (till March 16th, 2021). The proposed method is also easy to be plug-and-play in many cutting-edge 3D detection frameworks at negligible cost to boost performance. The code will be made publicly available.
\end{abstract}

\begin{figure}[t]
 \centering
  \includegraphics[width=0.95\linewidth]{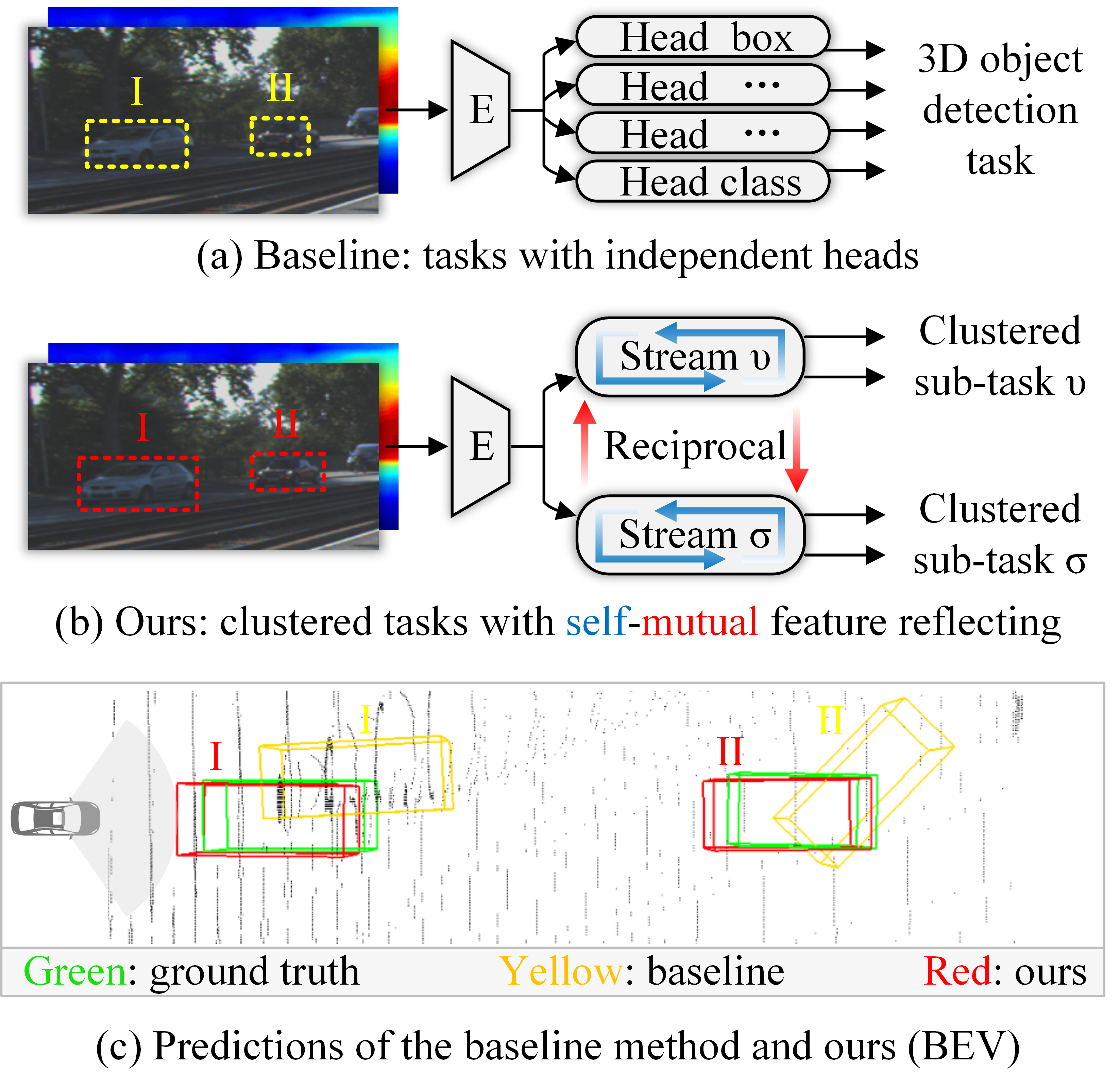}
 \caption{
 Comparison of the baseline method ($\rm D^4LCN$~\cite{ding2020learning}) and our proposed DFR-Net. 
 (a) The baseline method first uses the encoder (``E'') to extract RGB and predicted depth features and then adopts independent heads to decode the shared features for 3D detection tasks.
 (b) In our method, we first cluster the sub-tasks with common characteristics to build up separated task streams and then exploit the reciprocal information between the streams via self-mutual feature reflecting.
 (c) The ground truth, the prediction of the baseline method, and our prediction are shown in the bird's-eye view (BEV) pseudo LiDAR for better visualization.
}
 \label{fig1}
\end{figure}
\section{Introduction}
Building on the promising progress achieved by 2D object detection in recent years~\cite{ren2015faster, redmon2018yolov3}, vision- and LiDAR-based 3D object detection have received increasing attention from both industry and academia due to their critical roles in outdoor autonomous driving~\cite{geiger2012we} and indoor robotic navigation. 3D object detectors based on expensive LiDAR sensors~\cite{yan2018second, shi2019pointrcnn,9156880, 9511841} have been widely developed and excelled in 3D object detection, whereas a much cheaper alternative, i.e., monocular 3D object detection, remains an open and challenging research field.

Monocular 3D object detectors can be roughly divided into three categories according to different input data representations: RGB image-based, pseudo LiDAR-based, and depth-assisted image-based methods:
(i) RGB image-based methods aim to leverage geometry constraints~\cite{mousavian20173d} or semantic knowledge~\cite{chen2016monocular} to explore 2D-3D geometric consistency for recovering 3D location and dimension. However, the performance is still far from satisfactory due to the lack of reliable depth prior and the variance of the object scale caused by perspective projection. 
(ii) Pseudo LiDAR-based methods~\cite{wang2019pseudo,weng2019monocular} utilize depth estimation to reconstruct point clouds from image pixels. Afterwards cutting-edge LiDAR-based approaches such as \cite{qi2018frustum, shi2019pointrcnn} can be directly borrowed. Recent works~\cite{wang2019pseudo,weng2019monocular,you2019pseudo} have demonstrated the effectiveness of pseudo LiDAR-based methods. Nevertheless, due to the inaccurate depth prediction, a lack of RGB context as well as the inherent difference between real- and pseudo-LiDAR, the performance is limited. 
(iii) Depth-assisted methods such as~\cite{ding2020learning, shi2020distance} focus on the integration strategy of RGB and depth features, whereas the network is unable to resolve the inferior 3D localization due to the mis-estimated depth map. In other words, the performance relies heavily on the quality of depth maps.

Humans can get some hints about 3D information even from monocular cues
because the brain has the capability to utilize reciprocal visual information from different perception tasks \cite{du20203dcfs}, e.g., object localization and appearance perception (classification).
For example, if we know the category and size of an object, we will know how far away the object is.
On the other hand, if we know the localization and fuzzy scale of a distant or occluded unknown object, we may accordingly guess its category.

Inspired by humans' object perception system, we introduce a novel dynamic feature reflecting network for monocular 3D object detection, named DFR-Net. 
A novel appearance-localization feature reflecting module (ALFR) is designed, where 3D detection tasks are divided into two categories, the appearance perception tasks and the object localization tasks. Distinct tasks are sent to one of the two streams accordingly to delve into the task-specific features within each task, where reciprocal features between two categories self-mutually reflect.
Here the terminology ``reflect" denotes task-wise implicit feature awareness and correlation.
To further optimize the multi-task learning, we propose a dynamic intra-trading module, named DIT, which realigns the training process of two sub-tasks in a self-learning manner.
Figure \ref{fig1} shows the comparison of independent heads (baseline $\rm D^4LCN$~\cite{ding2020learning}) and our proposed DFR-Net. 
DFR-Net exploits and leverages reciprocal appearance-localization features for 3D reasoning and achieves superior performance. 
The proposed module is demonstrated to be effective on various image-based and depth-assisted image-based backbone networks (e.g., M3D-RPN~\cite{brazil2019m3d} and $\rm D^4LCN$~\cite{ding2020learning}). 

Our main contributions are summarized as follows: 
\vspace{-\topsep}
\begin{itemize}
\setlength{\parskip}{0pt}
\setlength{\itemsep}{0pt plus 1pt}
\item We introduce a simple yet effective dynamic feature reflecting network (DFR-Net) for monocular 3D object detection, which exploits the reciprocal information underlying the task, allowing the sub-tasks to benefit from each other to alleviate the ill-posed problem of monocular 3D perception.
\item We present an appearance-localization feature reflecting module (ALFR) that first separates two task streams and then self-mutually reflects the subtask-aware features.
\item We investigate a dynamic intra-trading module (DIT) that reweights different task losses to realign the multi-task training process in a self-learning manner.
\item We achieve new state-of-the-art monocular 3D object detection performance on the KITTI benchmark.
The method can be plug-and-play in many other frameworks to boost the performance at negligible cost.
\end{itemize}

\section{Related work}

\begin{figure*}[t]
 \centering
  \includegraphics[width=1.0\linewidth]{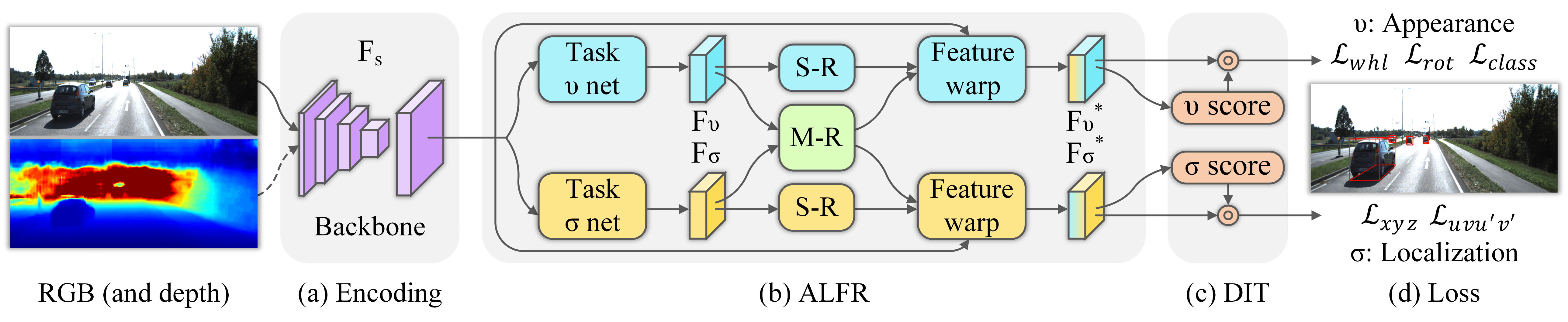}
 \caption{
  Schematic illustration of the proposed DFR-Net. (a) The RGB image and pseudo depth map (optional) are encoded as the shared feature $F_{s}$. (b) The proposed ALFR module first separates the feature into $F_{\upsilon}$ and $F_{\sigma}$ and then self-reflects (S-R) and mutual-reflects (M-R) the features and finally warps the features to get ${F_{\upsilon}}^*$ and ${F_{\sigma}}^*$. (c) the DIT module dynamically predicts scores from two warped features and realigns the training process according to the scores in a self-learning manner. (d) the losses for the task $\upsilon$ (appearance perception) and the task $\sigma$ (object localization).
}
 \label{fig2}
\end{figure*}

\noindent\textbf{Image-based detection}
Single image-based 3D object detection is much more challenging compared with stereo- and LiDAR-based detection because monocular 3D perception is an ill-posed problem, thus spatial information is not sufficient and precise enough for reliable object localization.
Some pioneering works~\cite{chen20153d,chen2016monocular,xu2018multi,qin2019monogrnet,atoum2017monocular,mousavian20173d,ku2019monocular,rubino20173d} attempted to use RGB and auxiliary information, e.g., semantic knowledge and geometry consistency, to alleviate this issue.
Brazil et al. introduced M3D-RPN~\cite{brazil2019m3d} that exploits the geometric relationship between 2D and 3D perspectives through sharing anchors and classification targets.
Inspired by the key point-based 2D object detector CenterNet~\cite{zhou2019objects}, RTM3D~\cite{RTM3D} estimates the nine projected key points of a 3D bounding box in image space to construct the geometric relationship of 3D and 2D to recover the 3D object information, whereas MonoPair~\cite{chen2020monopair} leverages spatial relationships between paired objects to effectively detect occluded objects.
Simonelli et al. proposed an alternative method MonoDIS~\cite{simonelli2019disentangling} that leverages a novel disentangling transformation for 2D and 3D detection losses, which conducts disentangled training of the losses from heterogeneous sets
of parameters to make the training process of monocular 3D detection more stable. However, this approach ignores to exploit and utilize the reciprocal information underlying the entire task.
Recently, researchers endeavor to adopt extra information to introduce much richer representations for the task. Monocular sequence-based Kinematic3D~\cite{brazil2020kinematic} efficiently utilizes 3D kinematic motion from videos to boost 3D detection performance. Reading et al. introduced CaDDN~\cite{CaDDN} that learns categorical depth distribution for each pixel in 2D space to project contextual information to the appropriate depth interval in 3D space.

\noindent\textbf{Pseudo depth-assisted detection}
Monocular depth estimation~\cite{fu2018deep} opens up an alternative and effective way for accurate monocular 3D object detection~\cite{manhardt2019roi}.
$\rm D^4LCN$ \cite{ding2020learning} introduced a local convolutional 3D object detection network, where depth maps were regarded as the guidance to learn local dynamic depthwise-dilated kernels for images, while the local dilated convolution can not fully capture the object context in the condition of perspective projection and occlusion. Different from these methods, pseudo LiDAR-based methods such as~\cite{wang2019pseudo, you2019pseudo} converting depth maps into artificial point clouds and adopting off-the-shelf LiDAR-based algorithms. 
Ma et al. observed that the efficacy of pseudo-LiDAR representation comes from the coordinate transformation and proposed PatchNet~\cite{Ma_2020_ECCV} that organizes the pseudo-LiDAR data as the image representation. However, most of the abovementioned methods are not fast enough for real-time applications.

\section{Methodology}

\subsection{Pipeline overview}
\label{Section:pipeline}
The goal of this work is to dig deep into the 3D detection problem and reformulate it as object localization and appearance perception sub-tasks
that can benefit from each other via reciprocal feature reflecting. 
%
Given a single image, 3D object detection aims to predict categories, 3D locations, dimensions as well as direction. Inspired by 2D object detection, categories of an instance can be inferred from the appearance features. Besides, since the dimensions of a certain object type usually have similar sizes, the rough dimension information can also be deduced from the appearance features. On the contrary, 3D locations vary along with the positions within the image. 
Motivated by this observation, we reformulate the task and propose the DFR-Net to divide the shared features into two siamese task-specific streams of localization and appearance, and exploit the intrinsic reciprocal information underlying the task to boost the performance at negligible cost.

As shown in Figure \ref{fig2}, the entire network is built upon an encoder-decoder architecture. 
For backbone network, our DFR-Net can employ various monocular 3D object detection approaches, such as M3D-RPN~\cite{brazil2019m3d} and $\rm D^4LCN$~\cite{ding2020learning}, etc. 
We adopt a depth-assisted monocular method~\cite{ding2020learning} to instantiate our model. Given the RGB image and estimated depth map, we fetch the shared features of the last convolutional layer of the encoder and feed them into two sub-task streams to explore the task-specific information. We design the self-mutual appearance-localization feature reflecting (ALFR) module to take a deep look into the implicit interaction across sub-tasks. 
In specific, two modules termed self-reflect module (S-R) and mutual-reflect module (M-R), are proposed. S-R is to delve into the task-specific features within each task whereas M-R combines the corresponding features between different tasks to diffuse and aggregate the mutual characteristics. 
Previous works \cite{brazil2019m3d,ding2020learning,RTM3D} usually leverage distinct head over the shared feature extracted by the encoder to regress disentangled targets. 
Differently, we learn appearance-dependent information (rotation, 3D dimension, category) from the appearance-aware stream and location-dependent knowledge (2D, 3D location) from the localization-aware stream. In order to realign the multi-task training process, we further design a DIT module to reweight different task losses for the joint optimization of each sub-task and thus contribute to the overall precision of 3D object perception.

\subsection{Appearance-localization feature reflecting}
\label{Section:ALFR}

\begin{figure}[htb]
 \centering
  \includegraphics[width=1.0\linewidth]{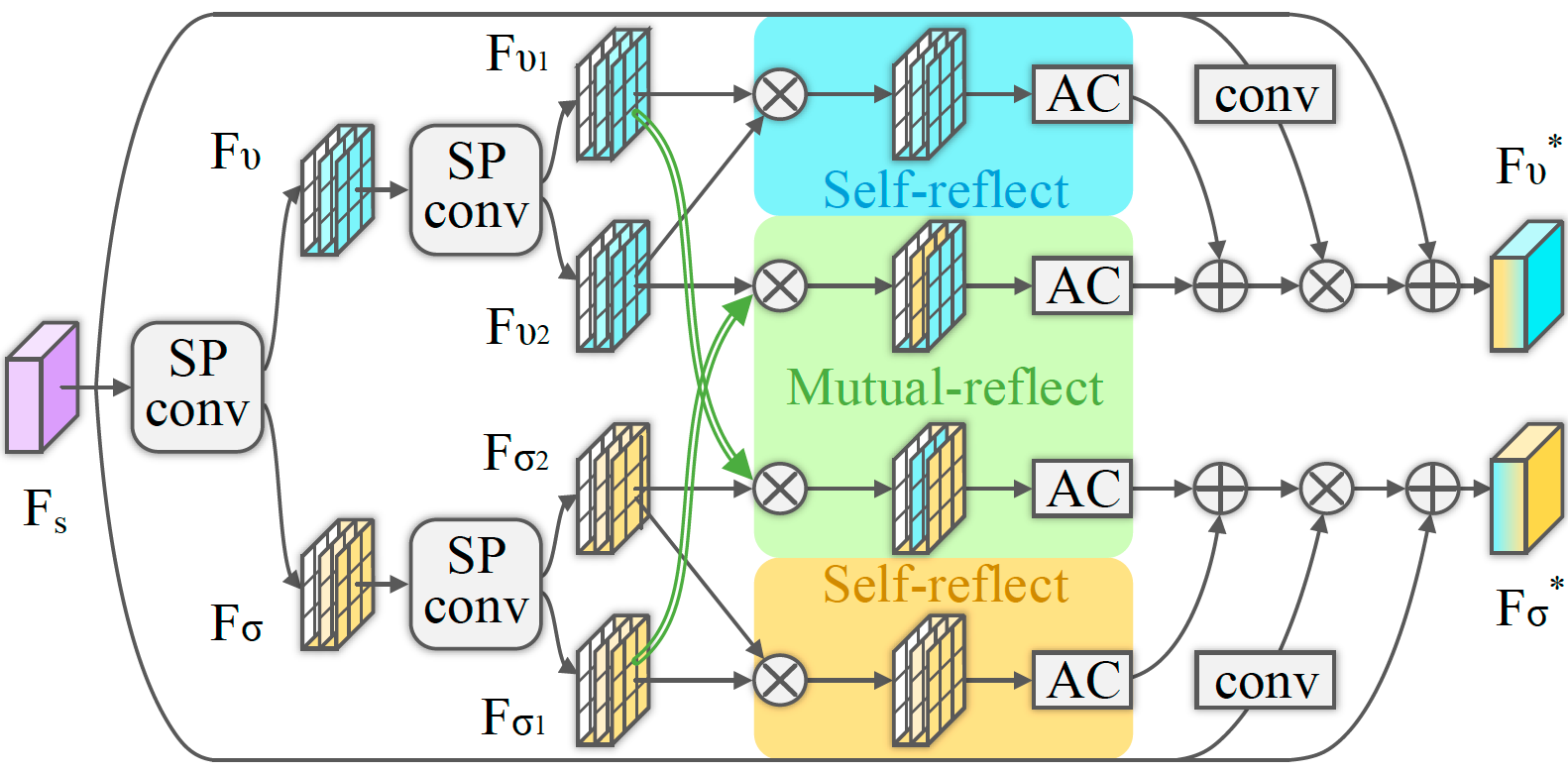}
 \caption{
  Illustration of the proposed ALFR. ``SP conv'' and ``AC'' denote the convolutions that separate the feature into different streams and the activation function, respectively. Self-reflect (S-R) and mutual-reflect (M-R) sub-modules are highlighted with different colors.
 }
 \label{fig3}
\end{figure}

To take a deep excavation of reciprocal information underlying monocular 3D object detection task, we propose an appearance-localization feature reflecting module (ALFR) to separate the shared feature into task-specific features and self-mutually reflect the reciprocal relations. As is shown in Figure \ref{fig3}, given a shared feature map $ F_s$, the module first applies two convolutional layers to generate two task-specific feature maps: appearance-specific feature $F_{\upsilon}$ and localization-specific feature $F_{\sigma}$. Then we feed $F_{\upsilon}$ into two convolution layers to generate two new feature maps $F_{\upsilon 1}$ and $F_{\upsilon 2}$, and meanwhile utilize $F_{\sigma}$ to generate another two new feature maps $F_{\sigma 1}$ and $F_{\sigma 2}$ in the same way. We design a self-reflect module (S-R) to capture the pair-wise context information within each task. 
Take the appearance stream in the upper part as an example (colored in blue), S-R takes context-aware $F_{\upsilon 1}$ and $F_{\upsilon 2}$ as inputs to calculate the self-reflect attention map of appearance $W_{\upsilon s}$. 
Besides, we feed $F_{\upsilon 2}$ and $F_{\sigma 1}$ into mutual-reflect (M-R) module to build mutual correlations across tasks and obtain the the mutual-reflect attention map of appearance $W_{\upsilon m}$. 
The self-reflect and mutual-reflect attention maps are combined via a learnable scale parameter to get the appearance-aware attention map $W_{\upsilon}$. 
To avoid the negative impact of noisy attention at the initial stage of the network, we design an adaptive residual connection between $W_{\upsilon m}$ and the shared input $F_s$ to get the final appearance-specific features $F_\upsilon^*$. 

In detail, the shared feature can be defined as $F_s \in \mathbb{R}^{C \times H \times W}$. Then the output feature maps are $\{F_{\upsilon 1},F_{\upsilon 2},F_{\sigma 1},F_{\sigma 2} \} \in \mathbb{R}^{C/r \times H \times W}$, where the reduction ratio $r$ is to reduce parameter overhead. We reshape them to $\mathbb{R}^{C/r \times N}$, where $N =H \times W$ represents the pixel number of each channel in features. In S-R, we perform a matrix multiplication between the transpose of $F_{\upsilon 1}$ and $F_{\upsilon 2}$ and apply a softmax layer to calculate the self-reflect attention map of appearance $W_{\upsilon s} \in \mathbb{R}^{N \times N}$:
\begin{equation}
    W_{\upsilon s} = \frac{exp((F_{\upsilon 1})^T \cdot F_{\upsilon 2})}{\sum_{j=1}^{N}exp((F_{\upsilon 1})^T \cdot F_{\upsilon 2})}
\end{equation}
In M-R, we perform a matrix multiplication between the transpose of $F_{\upsilon 2}$ and $F_{\sigma 1}$ and then apply a softmax layer to obtain the mutual-reflect attention map of appearance $W_{\upsilon m} \in \mathbb{R}^{N \times N}$, which can be formulated as:
\begin{equation}
    W_{\upsilon m} = \frac{exp((F_{\sigma 1})^T \cdot F_{\upsilon 2})}{\sum_{j=1}^{N}exp((F_{\sigma 1})^T \cdot F_{\upsilon 2})}
\end{equation}
Then we perform an element-wise sum operation to combine the self-reflect and mutual reflect attention maps and output the appearance-aware attention maps $W_{\upsilon}$ as:
\begin{equation}
    W_{\upsilon} = \lambda_\upsilon * W_{\upsilon s} + (1 - \lambda_\upsilon) * W_{\upsilon m}
\end{equation}
We send the share feature into a convolution layer to generate a new feature map $F_{\upsilon s}$ and reshape it from $\mathbb{R}^{C \times H \times W}$ to $\mathbb{R}^{C \times N}$ and perform matrix multiplication between $F_{\upsilon s}$ and the transpose of $W_{\upsilon}$:
\begin{equation}
    F'_{\upsilon s} = F_{\upsilon s} \cdot (W_{\upsilon})^T
\end{equation}
Finally, we reshape $F'_{\upsilon s} \in \mathbb{R}^{C \times N}$ to $\mathbb{R}^{C \times H \times W}$ and combine it with the shared feature $F_s$ via a learnable parameter $\beta_\upsilon$ to get the final appearance-specific features $F_\upsilon^*$:
\begin{equation}
    F_\upsilon^*  = F_s + \beta_\upsilon * F'_{\upsilon s}
\end{equation}
The final localization-specific $F_\sigma^*$ is generated similarly.

\subsection{Dynamic intra-trading}
\label{Section:DIT}

\begin{figure}[htb]
 \centering
  \includegraphics[width=1.0\linewidth]{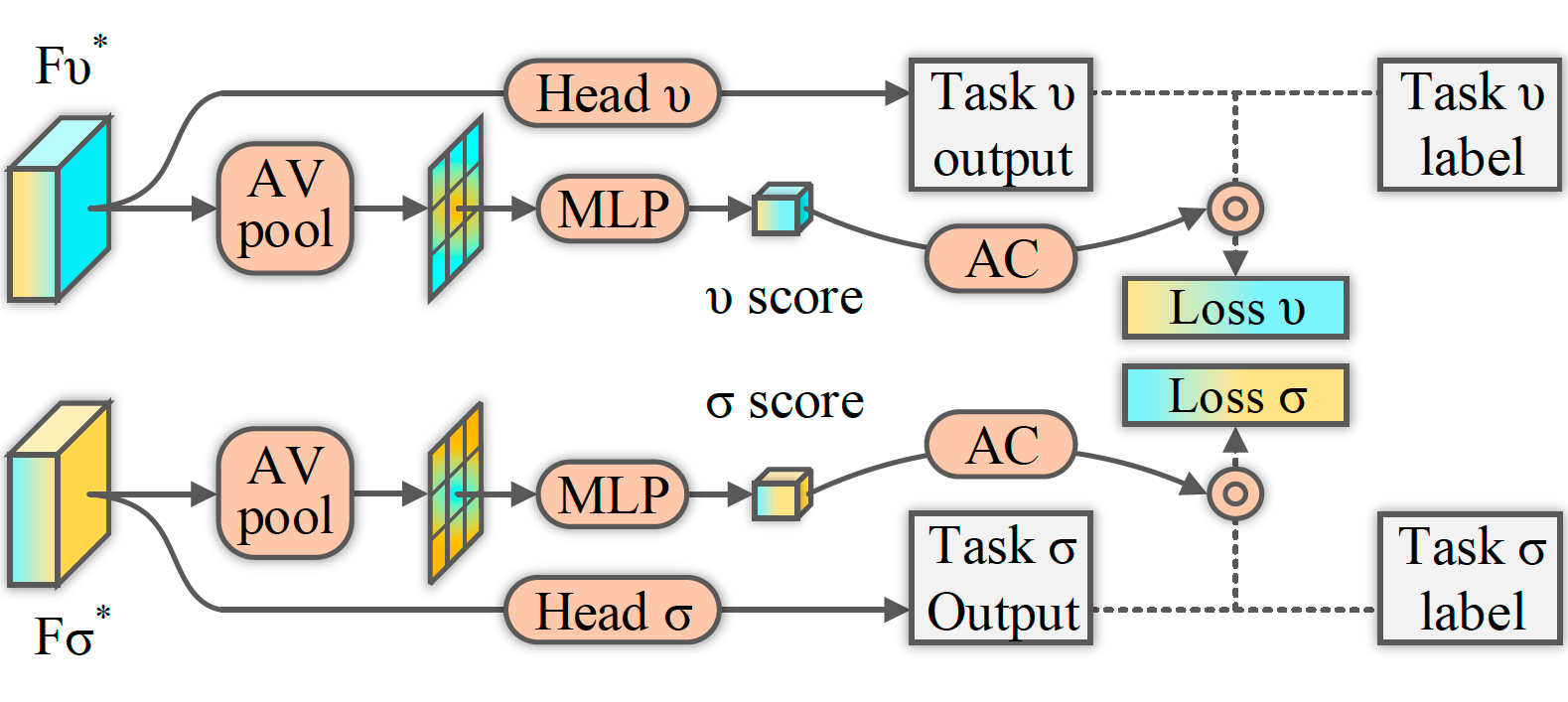}
 \caption{
  Illustration of the proposed DIT. ``AV pool'' denotes the average pooling operation. ``AC'' represents the activation function. The scores that realigns the training processes of tasks $\upsilon$ and $\sigma$ are generated from $F_\upsilon^*$ and $F_\sigma^*$, respectively. The losses of sub-tasks are reweighted via the self-learned scores.
}
 \label{fig3-2}
\end{figure}

Thanks to the ALFR module, we can generate appearance-related output ($\{class,w,h,l,rot\}$) from the appearance-aware feature $F_\upsilon^*$ and localization-related output ($\{u,v,u',v',x,y,z\}$) from the localization-aware feature $F_\sigma^*$. To measure the differences between the outputs and the ground truth, we define the the appearance-aware loss $L_\upsilon$ for the task $\upsilon$ and the localization-aware loss $L_\sigma$ for the task $\sigma$. 
The naive mode to combine these two losses could be to conduct a linear sum directly, but model performance will be badly affected by the choice of the hyper-parameters weights for each task loss. 
To enhance the joint optimization of each task, we propose a novel dynamic intra-trading module (DIT) to adaptively learn the confidence scores depending on the actual contribution of the task-related losses for the joint learning. 
In specific, we first sent the appearance-aware features $F_\upsilon$ to the average pooling layer to aggregate the context information at the spatial level. Then the outputs go through the multilayer perceptron (MLP) and a Sigmoid layer to get the appearance-aware trading score $S_{\upsilon}$, which indicates the semantic confidence contained in the input sample. In the same way, we utilize the localization-aware features $F_{\sigma}$ to obtain the localization-aware trading score $S_\sigma$. These two scores are used to guide the network learning by weighting the task-related losses. The total trading loss can be formulated as:
\begin{equation}
\label{equ_total}
    L = S_\upsilon * L_\upsilon + S_\sigma * L_\sigma - log(S_\upsilon * S_\sigma)
\end{equation} 
where $log(S_\upsilon * L_\upsilon)$ is the regularization term for the training weights. When $S_\upsilon$ or $S_\sigma$ increases,  $log(S_\upsilon * L_\upsilon)$ increases to maintain the overall balance of the total losses and vice versa. Moreover, the appearance-aware trading score $S_{\upsilon}$ will turn small when the input samples are not confident in semantic information. The same applies to the localization-aware trading score $S_\sigma$. This strategy avoids the negative effect of much noise on the network optimization. In this way, the networks can adjust the proportion of loss back-propagation during the training process and thus promote the learning accuracy of two tasks.

\subsection{Objective functions}
\label{Section:losses}
Following the baseline method~\cite{brazil2019m3d}, we define the losses of each group as:
\begin{equation}
    L_{class} = -log(\frac{exp(class_t)}{\sum_{j}^{n_c}{exp(class_j)}})
\end{equation}
\begin{equation}
    L_{rot} = SmoothL1([rot], [{rot}_g])  
\end{equation}
\begin{equation}
    L_{whl} = SmoothL1([w,h,l], [w_g,h_g,l_g])  
\end{equation}
\begin{equation}
    L_{uvu'v'} = -log( IOU([uvu'v'],[u_gv_gu'_gv'_g]))
\end{equation}
\begin{equation}
    L_{xyz} = SmoothL1([x,y,z],[x_g,y_g,z_g])
\end{equation}
where $n_c$ indicates the number of the categories in the training set. We use the standard cross-entropy (CE) for the classification loss and the Smooth L1 for other regression losses. 
The appearance-aware and the localization-aware losses are formulated as follows:
\begin{equation}
    L_\upsilon = L_{class} + L_{rot} + L_{whl}
\end{equation}
\begin{equation}
    L_\sigma =  L_{uvu'v'} + L_{xyz}
\end{equation}
The total loss is defined in Equation \ref{equ_total}.
\section{Experiments}

\begin{table*}[t]
 \small
 \begin{center}
 \begin{tabular}{|C{2.6cm}|C{2.2cm}|C{0.95cm}|C{0.95cm}|ccc|ccc|c|}
  \hline
  \multirow{2}*{Method} & \multirow{2}*{Reference} & Speed & Extra & \multicolumn{3}{c|}{$\rm AP_{3D}$} & \multicolumn{3}{c|} {$\rm AP_{BEV}$} & \multirow{2} * {GPU} \\
                        &                          & (FPS) & Info. & Mod.    & Easy     & Hard          & Mod.    & Easy     & Hard            &                      \\
  \hline
  \hline
    FQNet\cite{liu2019deep}                  &CVPR 2019           & 2    & -          & 1.51  & 2.77  & 1.01  & 3.23  & 5.40  & 2.46  & 1080Ti     \\
   
    MonoGRNet\cite{qin2019monogrnet}         &AAAI 2019           & 16   & -          & 5.74  & 9.61  & 4.25  & 11.17  & 18.19  & 8.73  & Tesla P40  \\
    MonoDIS\cite{simonelli2019disentangling} &ICCV 2019           & -    & -          & 7.94  & 10.37 & 6.40  & 13.19 & 17.23 & 11.12 & Tesla V100 \\
    MonoPair\cite{chen2020monopair}          &CVPR 2020           & 17   & -          & 9.99  & 13.04 & 8.65  & 14.83 & 19.28 & 12.89 & -          \\
    UR3D \cite{shi2020distance}              &ECCV 2020           & 8    & -          & 8.61  & 15.58 & 6.00  & 12.51 & 21.85 & 9.2   & GTX Titan X\\
    \rowcolor{maroon!10} M3D-RPN 
    \cite{brazil2019m3d}                     &ICCV 2019           & 6.2  & -          & 9.71  &	14.76 &	7.42  &	13.67 & 21.02 &	10.23 & 1080Ti     \\
    RTM3D\cite{RTM3D}                        &ECCV 2020           & 20   & -          & 10.34 & 14.41 & 8.77  & 14.20 & 19.17 & 11.99 & 1080Ti     \\
    \rowcolor{maroon!10}DFR-Net (I)   & -    & 6.1  & -          & \textbf{11.89}    & \textbf{17.30}      & \textbf{9.32}      & \textbf{16.47}     & \textbf{24.38}     & \textbf{13.33}     & 1080Ti     \\ 
    \hline
    \hline
    AM3D\cite{ma2019accurate}                &ICCV 2019           & 3    & Depth      & 10.74 & 16.50 & 9.52  & 17.32 & 25.03 & 14.91 & 1080Ti     \\
    PatchNet\cite{Ma_2020_ECCV}              &ECCV 2020           & 3    & Depth      & 11.12 & 15.68 & 10.17 & 16.86 & 22.97 & 14.97 & 1080       \\
    DA-3Ddet\cite{Ye_2020_ECCV}              &ECCV 2020           & 3    & D + L      & 11.50 & 16.77 & 8.93  & 15.90 & 23.35 & 12.11 & Titan RTX  \\
    \rowcolor{myblue}
    $\rm D^4LCN$\cite{ding2020learning}      &CVPR 2020           & 5.6  & Depth      & 11.72 & 16.65 & 9.51  & 16.02 & 22.51 & 12.55 & 1080Ti     \\
    Kinematic3D\cite{brazil2020kinematic}    &ECCV 2020           & 8    & Video      & 12.72 & 19.07 & 9.17  & 17.52 & 26.69 & 13.10 & -           \\
    CaDDN \cite{CaDDN}                       &CVPR 2021           & 2    & LiDAR      & 13.41 & 19.17 & \textbf{11.46}& 18.91 & 27.94 & \textbf{17.19}    
                                                                                                                                      & Tesla V100  \\
    \rowcolor{myblue}DFR-Net (I+D)                          & -                  & 5.5    & Depth      & \textbf{13.63} & \textbf{19.40} & 10.35 
                                                                                                           & \textbf{19.17} & \textbf{28.17} & 14.84  & 1080Ti \\ 
  \hline
 \end{tabular}
 \end{center}
  \caption{
  Comparison with state-of-the-art (SoTA) methods on the KITTI test set at IoU = 0.7 (R40).
 ``Depth'' and ``Video'' denote using prior depth estimation and video sequence as an extra input, respectively.
 ``LiDAR'' denotes using LiDAR point clouds as extra supervision.
 ``D + L'' denotes using both ``Depth'' and ``LiDAR''.
 Based on the encoding backbone of M3D-RPN~\cite{brazil2019m3d} (the pink row), we rank $1^{st}$ among all the image-only-based methods. Based on the backbone of $\rm D^4LCN$~\cite{ding2020learning} (the cyan row), we rank $1^{st}$ among all the competitors in the KITTI monocular 3D object detection track with a high inference speed (2$\times$ faster than the second even with a much lighter GPU).
 }
 \label{table_kitti_test}
\end{table*}

\noindent\textbf{Dataset}
Experiments are conducted on the challenging KITTI dataset~\cite{geiger2013vision, geiger2012we}, which contains 7,481 and 7,518 images for training and testing, respectively. Following previous works~\cite{brazil2019m3d, ding2020learning}, we utilize two train-val splits: ``val1'' split contains 3,712 training and 3,769 validation images while ``val2'' split employs 3,682 images for training and 3,799 images for validation. We comprehensively analyze the performance of the proposed DFR-Net with other methods on the test and two validation sets.

\noindent\textbf{Evaluation metrics}
For evaluation, we use precision-recall curves and report the average precision (AP) performance of bird’s eye view (BEV) detection and 3D object detection on the KITTI validation and test set. The KITTI test server uses the 40 recall positions-based metric (R40) instead of the 11 recall positions-based metric (R11) after Aug. 2019. We denote AP for 3D and BEV detection as $\rm AP_{3D}$ and $\rm AP_{BEV}$, respectively. In the benchmark, three levels of difficulty are defined according to the 2D bounding box height, occlusion, and truncation degree, namely, ``Easy'', ``Mod.'', and ``Hard'', and the KITTI benchmark ranks all approaches based on the $\rm AP_{3D}$ of ``Mod.''. Following previous methods~\cite{chen2017multi, ding2020learning}, IoU = 0.7 is adopted as the threshold for the ``Car'' category, and IoU = 0.5 is adopted as the threshold for the ``Cyclist'' and ``Pedestrian'' categories.

\noindent\textbf{Training details}
Our experimental settings are strictly consistent with our image-based and depth-assisted baseline methods~\cite{brazil2019m3d,ding2020learning} for fair comparison. For M3D-RPN~\cite{brazil2019m3d}, we use a single Nvidia Tesla v100 GPU to train the model for 50k iterations. The learning rate is set to 0.004 with a poly rate using power as 0.9. We use a batch size of 2 and a weight decay of 0.9. For $\rm D^4LCN$~\cite{ding2020learning}, the network is optimized by SGD with a momentum of 0.9 and a weight decay of 0.0005. We use 4 Nvidia Tesla v100 GPUs to train the model for 40k iterations. The base learning rate is set to 0.01 and power to 0.9. For both methods, the input images are scaled to 512$\times$1760, and horizontal flipping is the only data augmentation. The reduction ratio $r$ is set to 8. During inference, we apply non-maximum suppression (NMS) on the box outputs in the 2D space using IoU criteria of 0.4 and filter boxes with scores below 0.75.

\subsection{Comparison with state-of-the-arts}
\noindent\textbf{Results on the KITTI test set}
We first report the 3D ``Car'' detection results on KITTI test set at IoU = 0.7 in Table \ref{table_kitti_test}.
Our plug-and-play DFR-Net has two versions: (a) DFR-Net (I): the image-only-based model based on the backbone of M3D-RPN~\cite{brazil2019m3d} (the pink row); (b) DFR-Net (I+D): the depth-assisted model based on the backbone of $\rm D^4LCN$~\cite{ding2020learning} (the cyan row).
In the KITTI leaderboard, our DFR-Net (I+D) ranked $1^{st}$ among all the monocular-based 3D object detection methods. 
Note that among all the image-only-based detectors, our DFR-Net (I) still ranks $1^{st}$ and outperforms them with a considerable margin. 

Compared with Kinematic3D~\cite{brazil2020kinematic} that utilizes multiple frames to leverage the temporal motion information to boost the performance, our method achieves superior performance with an improvement of ( 0.33\% / 0.91\% / 1.18\% ) on ``Easy'', ``Mod.'', and ``Hard'', respectively. Compared with the previous top-ranked CaDNN~\cite{CaDDN}, our method still carries out superior results on ``Easy'' and ``Mod.'' and a comparable result on ``Hard''. Note that the proposed DFR-Net (I+D) can get a real-time speed of 40 FPS on Tesla V100, which is 20 times faster than CaDNN~\cite{CaDDN}. The proposed ALFR module only occupies a small computational cost. Therefore, the inference speed and model size are comparable with the baseline method $\rm D^4LCN$ (5.5 vs. 5.6 FPS; 355 vs. 352 Mb).

\begin{figure*}[htb]
 \centering
 \includegraphics[width=17.4cm]{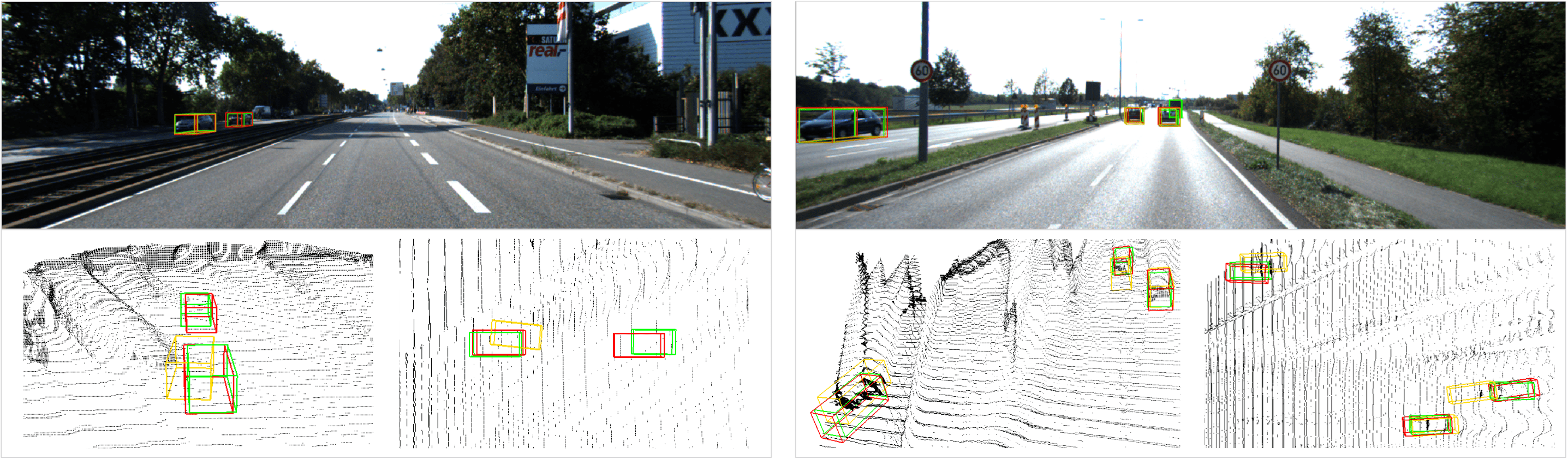}
 \caption{
 The qualitative comparison of the ground truth (green), the baseline (yellow), and our method (red) on the KITTI validation set. For better visualization, the first row shows RGB images, and the second row shows the front view (left) and BEV (right) pseudo LiDAR, respectively. Due to the reciprocal feature reflecting, the proposed approach can predict accurate 3D bounding boxes of distant objects even with inaccurate depth estimation.
 }
 \label{fig4}
\end{figure*}

\begin{table}[t]
 \small
 \begin{center}
 \begin{tabular}{|C{2.0cm}|C{0.6cm}C{0.6cm}C{0.6cm}|C{0.6cm}C{0.6cm}C{0.6cm}|}
  \hline
  \multirow{2} * {Method}                      & \multicolumn{3}{c|}{$\rm val1$}       & \multicolumn{3}{c|} {$\rm val2$}  \\
                                               & Easy     & Mod.      & Hard                      & Easy     & Mod.     & Hard                                       \\
  \hline
  \hline
 M3D-RPN \cite{brazil2019m3d}             & 14.53 & 11.07 & 8.65 & 14.57 & 10.07 & 7.51 \\
 Ours                                     & \textbf{19.55} & \textbf{14.79} & \textbf{11.04} & \textbf{19.38} & \textbf{14.33} & \textbf{10.63} \\
 \rowcolor{maroon!10}\textit{Improvement} & \textit{+5.02} & \textit{+3.72} & \textit{+2.39} & \textit{+4.81} & \textit{+4.26} & \textit{+3.12} \\
 \hline
 $\rm D^4LCN$ \cite{ding2020learning}            & 22.32 & 16.20 &  12.30 & 22.07 & 14.41 & 10.39 \\
 Ours                        &  \textbf{24.81}  & \textbf{17.78} & \textbf{14.41} & \textbf{24.30} & \textbf{17.23} &  \textbf{12.52} \\
 \rowcolor{myblue}\textit{Improvement}                       & \textit{+2.49} & \textit{+1.58} & \textit{+2.11} & \textit{+2.23} & \textit{+2.82} & \textit{+2.13} \\
  \hline
 \end{tabular}
 \end{center}
  \caption{$\rm AP_{3D}$  performance for the ``Car'' category on KITTI ``val1'' and ``val2'' split set at IoU = 0.7 (R40).}
  \label{val12}
\end{table}

\begin{table}[t]
 \small
 \begin{center}
 \begin{tabular}{|C{2.0cm}|C{0.6cm}C{0.6cm}C{0.6cm}|C{0.6cm}C{0.6cm}C{0.6cm}|}
  \hline
  \multirow{2} * {Method}                      & \multicolumn{3}{c|}{Pedestrian}                   & \multicolumn{3}{c|} {Cyclist}                                   \\
                                               & Easy     & Mod.      & Hard                      & Easy     & Mod.     & Hard                  \\
  \hline
  \hline
  M3D-RPN \cite{brazil2019m3d}& 4.92  & 3.48 & 2.94 & 0.94 & 0.65 & 0.47 \\
  Ours  & \textbf{6.62}  & \textbf{4.58} & \textbf{4.17} 
        & \textbf{1.63} & \textbf{1.01} & \textbf{1.02} \\
  \rowcolor{maroon!10}\textit{Improvement} & \textit{+1.70} & \textit{+1.10} & \textit{+1.23} & \textit{+0.69} & \textit{+0.36} & \textit{+0.55} \\
  \hline
  $\rm D^4LCN$\cite{ding2020learning}       & 4.55           & 3.42           &2.83            & 2.45           & 1.67           & 1.36          \\
  Ours                                      & \textbf{6.09}  & \textbf{3.62}  & \textbf{3.39}  & \textbf{5.69}  & \textbf{3.58}  & \textbf{3.10} \\ 
  \rowcolor{myblue}\textit{Improvement}     & \textit{+1.54} & \textit{+0.20} & \textit{+0.56} & \textit{+3.24} & \textit{+1.91} & \textit{+1.74} \\
  \hline
 \end{tabular}
 \end{center}
  \caption{$\rm AP_{3D}$ performance for ``Cyclist'' and ``Pedestrian'' on KITTI test set at IoU = 0.5 (R40).}
  \label{class}
\end{table}

\noindent\textbf{Results on the KITTI validation set}
We evaluate the proposed framework compared with the cutting-edge image-based~\cite{brazil2019m3d} and depth-assisted~\cite{ding2020learning} baseline methods on the ``val1'' and ``val2'' split sets using $\rm AP_{40}$ as the evaluation metric, as shown in Table \ref{val12}. 
Since $\rm D^4LCN$~\cite{ding2020learning} report only the result of ``val1'' split, we reproduce the results by using the official public code. The proposed method improves the overall accuracy by a large margin compared to the baseline. For instance, the $\rm AP_{3D}$ performance of M3D-RPN~\cite{brazil2019m3d} on ``val1'' set gains ( 5.02\% / 3.72\% / 2.39\% ) improvement with the contribution of our DFR-Net on ``Easy'', ``Mod.'', and ``Hard'', respectively. Qualitative comparisons of the baseline and our method are shown in Figure~\ref{fig4}. The ground truth, baseline, and our method are colored in green, yellow, and red, respectively. For better visualization, the first and second columns show RGB images and BEV images of pseudo point clouds, respectively. Compared with the baseline, our DFR-Net can produce higher-quality 3D bounding boxes in different scenes. More quantitative and qualitative results are reported in our supplementary material.

\noindent\textbf{Results on ``Cyclist'' and ``Pedestrian''}
Due to the non-rigid structures and small scale of ``Cyclist'' and ``Pedestrian'' categories, it is much more challenging to detect these two categories. Pseudo-LiDAR based methods such as PatchNet \cite{Ma_2020_ECCV} and DA-3Ddet \cite{Ye_2020_ECCV} fail to detect ``Cyclist'' and ``Pedestrian''. 
We report these two categories with respect to two baseline methods \cite{brazil2019m3d,ding2020learning} in Table~\ref{class}. 
Following \cite{brazil2019m3d, ding2020learning}, $\rm AP_{3D}$ of ``Cyclist'' and ``Pedestrian'' on the test set at IoU = 0.5 (R40) are reported. Thanks to the reciprocal information underlying the task for 3D reasoning, we are able to localize these challenging categories to some degree and consistently outperform the baselines.

\subsection{Ablation study}
In this section, we choose M3D-RPN~\cite{brazil2019m3d} as the baseline and all experiments are conducted on KITTI ``val1'' split set.

\noindent\textbf{Main ablative analysis}
\label{ablation}
The DFR-Net consists of two modules: the ALFR and DIT modules. 
The ALFR module includes two sub-modules: the self-reflect (S-R) and mutual-reflect (M-R). To demonstrate the effectiveness of each module, we experiment with different combinations of sub-modules and the results are shown in Table \ref{ablation}. We can observe that the performance continues to grow with the participation of components. Specifically, from the comparison of group I and group II or group III and group IV, we find that adding the S-R module contributes to the model, improving the $\rm AP_{3D}$ (R40) performance on ``Mod.'' from 11.07\% to 13.08\% and 13.01\% to 13.39\%, respectively. The same goes for the M-R module. Meanwhile, combining two modules (S-R and M-R) together works better than merely using one of them separately. This can be concluded from the results of groups II, III, and IV, where the $\rm AP_{3D}$ (R40) performance on ``Mod.'' attains better performance 13.39\% compared to 13.08\% or 13.01\%. The above conclusions prove the effectiveness of our ALFR module. When simultaneously embedding the DIT module into the networks, the proposed model attains the best performance regardless of $\rm AP_{3D}$ or $\rm AP_{BEV}$ metric, which validates the effectiveness of the DIT module. 


\begin{table*}[htb]
 \small
 \begin{center}
 \begin{tabular}{|c|C{0.7cm}|C{0.7cm}|C{0.7cm}|ccc|ccc|}
  \hline
  \multirow{2} * {Group} & \multirow{2} * {S-R} & \multirow{2} * {M-R} &\multirow{2} * {DIT} & \multicolumn{3}{c|}{$\rm AP_{3D}$ (R11 / R40)} & \multicolumn{3}{c|} {$\rm AP_{BEV}$ (R11 / R40)}                             \\
       &            &            &            & Easy         & Mod.         & Hard         & Easy         & Mod.         & Hard       \\
  \hline
  \hline
  I    & -          &-           &-           & 20.27 / 14.53  & 17.06 / 11.07  & 15.21 / 8.65   & 25.94 / 20.85  & 21.18 / 15.62  & 17.90 / 11.88 \\
  II   & \checkmark &-           &-           & 20.75 / 17.30  & 16.57 / 13.08  & 14.98 / 10.41  & 27.62 / 24.61  & 22.65 / 18.03  & 18.50 / 14.61 \\
  III  & -          & \checkmark &-           & 20.27 / 17.26  & 17.11 / 13.01  & 14.36 / 10.50  & 25.09 / 23.26  & 21.65 / 17.79  & 17.47 / 13.82 \\
  IV   & \checkmark & \checkmark &-           & 21.08 / 18.00  & 17.10 / 13.39  & 15.19 / 10.78  & 26.53 / 24.87  & 22.01 / 18.35  & 17.66 / 15.03 \\
  V    & \checkmark &-           & \checkmark & 21.23 / 17.56  & 17.13 / 13.46  & 15.23 / 10.79  & 27.57 / 24.56  & 22.70 / 18.46  & 18.45 / 15.15 \\
  VI   & -          & \checkmark & \checkmark & \textbf{22.81} / 18.06  & 18.15 / 13.88  & 16.10 / 10.23  & 27.64 / 24.85  & 23.07 / 18.60  & 19.01 / 14.35 \\
  VII  & \checkmark & \checkmark & \checkmark & 22.04 / \textbf{19.55} & \textbf{18.43} / \textbf{14.79} & \textbf{16.96} / \textbf{11.04} & 
                                                \textbf{28.63} / \textbf{26.60} & \textbf{23.15} / \textbf{19.80} & \textbf{19.31} / \textbf{15.34} \\
  \hline
 \end{tabular}
 \end{center}
  \caption{Ablative analysis on the ``Car'' category on KITTI ``val1'' split set for $\rm AP_{3D}$ and $\rm AP_{BEV}$ at IoU = 0.7}
   \label{ablation}
\end{table*}

\begin{table}[htb]
 \small
 \begin{center}
 \begin{tabular}{|C{0.88cm}|C{0.88cm}|C{0.88cm}|C{0.88cm}|C{0.6cm}C{0.6cm}C{0.6cm}|}
  \hline
  \multicolumn{2}{|c|}{Task $\upsilon$}                & \multicolumn{2}{c|}{Task $\sigma$}      & \multicolumn{3}{c|}{$\rm AP_{3D}$}           \\
   \multicolumn{2}{|c|}{Loc: xyz, uvu'v'}              & \multicolumn{2}{c|}{App: class}         & Easy         & Mod.          & Hard          \\
  \hline
  \hline
  rotation & whl      & -           & -          & 17.21          & 13.35          & 10.73                  \\
  rotation & -        & -           & whl        & 18.55          & 14.14          & \textbf{11.29}         \\
  -        & whl      & rotation    & -          & 17.62          & 14.31          & 10.88                  \\
  -        & -        & rotation    & whl        & \textbf{19.55} & \textbf{14.79} & 11.04                  \\
  \hline
 \end{tabular}
 \end{center}
  \caption{$\rm AP_{3D}$ and $\rm AP_{BEV}$ comparison of different task clustering strategies on ``val1'' split set at IoU = 0.7 (R40).}
  \label{task}
\end{table}

\noindent\textbf{Different strategies of task clustering}
We conduct an incisive analysis on the effect of different task clustering strategies. The results are shown in Table \ref{task}. Since the task partition of some element variables is relatively certain, such as $\{x,y,z,u,v,u',v'\}$ belongs to object localization task (``Loc'') while $class$ remains with appearance perception task (``App''), we focus on the attribution of rotation and dimension $\{w,h,l\}$. The first and second row results demonstrate that 
$\{w,h,l\}$ clustered by the appearance task achieves a better performance of (18.55\% / 14.14\% / 11.29\%) than clustered by the localization task (17.21\% / 13.35\% / 10.73\%).
The first and third row results reveal that assigning the ``rotation'' to the appearance task attains a 0.96\% gain on ``Mod.'' compared to assigning it to localization task. When simultaneously allocating $\{w,h,l\}$ and rotation to the appearance-aware task achieves the best performance, which proves the effectiveness of our choice from the perspective of experiments.

\noindent\textbf{Information flow in the ALFR} 
We further dig into the information flow in the M-R module of our ALFR and experiment on the ``val1'' split set. Table~\ref{flow} reports the final performance of different forms of information flow. The M-R module is composed of two information flow: appearance to localization (``App$\rightarrow$~Loc'') and localization to appearance (``Loc$\rightarrow$~App''). Note that ``None'' denotes the DFR-Net without the M-R module (group V in Table~\ref{ablation}). From the table, we can find that adding one of these types of information can be beneficial. In specific, adding appearance to localization flow improves the ``Mod.'' performance from 13.46\% to 13.64\%, while adding localization to appearance flow promotes to 13.73\%. The combination of two flows achieves the best accuracy.

\begin{table}[htb]
 \small
 \begin{center}
 \begin{tabular}{|C{1.8cm}|C{0.6cm}C{0.6cm}C{0.6cm}|C{0.6cm}C{0.6cm}C{0.6cm}|}
  \hline
   \multirow{2} * {Method}    & \multicolumn{3}{c|}{$\rm AP_{3D}$}                   & \multicolumn{3}{c|} {$\rm AP_{BEV}$}         \\
                              & Easy           & Mod.           & Hard           & Easy           & Mod.        & Hard              \\
  \hline
  \hline
   None                       & 17.56          & 13.46          & 10.79          & 24.56          & 18.46          & 15.15           \\
   App $\rightarrow$ ~Loc     & 18.59          & 13.64          & 10.98          & 24.89          & 18.92          & 15.11           \\
   Loc $\rightarrow$ ~App     & 18.67          & 13.73          & \textbf{11.14} & 25.77          & 19.23          & \textbf{15.82}  \\
   Ours                       & \textbf{19.55} & \textbf{14.79} & 11.04          & \textbf{26.60} & \textbf{19.80} & 15.34           \\
  \hline
 \end{tabular}
 \end{center}
  \caption{$\rm AP_{3D}$ and $\rm AP_{BEV}$ comparison of different feature reflecting strategies on ``val1'' split set at IoU = 0.7 (R40).}
  \label{flow}
\end{table}

\noindent\textbf{Different settings of the DIT} In order to further dig into the impact of the DIT module, we define some variants of the DIT module based on different settings: 
(a) ``DIT-init'': initialize the trading scores for each task instead of network generation; 
(b) ``DIT-cross'': appearance-aware task and localization-aware task generate each other's trading scores; 
(c) ``DIT-shared'': the trading scores of each task are generated from the shared feature. 
As shown in Table \ref{DIT}, DIT-init improves the $\rm AP_{3D}$ performance from 13.39\% to 14.07\% in Mod. setting. However, it is obvious that DIT-cross distinctly drops the overall accuracy. This is because the final outputs of each task stream are specific to the corresponding task after the encoding of the ALFR module. Thus, there will be a lot of noise in predicting the confidence for another task, which affects the network learning.
DIT-shared achieves better results than the above designs, reaching to $\rm AP_{3D}$ 19.56\% on ``Easy'', which explains the shared-features contain the rich contextual information required by the two tasks. When equipped with the proposed DIT, the model can get the best performance, which demonstrates the effectiveness of our module.

\begin{table}[t]

 \small
 \begin{center}
 \begin{tabular}{|C{1.5cm}|C{0.6cm}C{0.6cm}C{0.6cm}|C{0.6cm}C{0.6cm}C{0.6cm}|}
  \hline
   \multirow{2} * {Method}    & \multicolumn{3}{c|}{$\rm AP_{3D}$}                   & \multicolumn{3}{c|} {$\rm AP_{BEV}$}             \\
                                  & Easy           & Mod.           & Hard           & Easy           & Mod.        & Hard              \\
  \hline
  \hline
   None                           & 18.00          & 13.39          & 10.78          & 24.87          & 18.35       & 15.03             \\
   DIT-init                       & 18.67          & 14.07          & 10.50          & 25.43          & 19.17       & 14.07             \\
   DIT-cross                      & 15.59          & 12.44          & 9.45           & 22.95          & 17.74       & 14.57             \\
   DIT-shared                     & \textbf{19.56} & 13.87          & 10.91          & 26.20          & 19.66       & 15.31             \\
   Ours                           & 19.55 & \textbf{14.79} & \textbf{11.04} & \textbf{26.60} & \textbf{19.80} & \textbf{15.34}          \\
  \hline
 \end{tabular}
 \end{center}
  \caption{$\rm AP_{3D}$ and $\rm AP_{BEV}$ comparison of different dynamic intra-trading methods on ``val1'' split set at IoU = 0.7 (R40).}
  \label{DIT}
\end{table}

\noindent\textbf{Generalization ability} For generalization ability validation, we extend our method to 2D detection task. We choose the well known SSD \cite{liu2016ssd} as the baseline and apply the proposed ALFR and DIT module for comparison. As illustrated in Table \ref{2D_detection}, we perform experiments on the VOC dataset in 07++12 setting: training on the union of VOC2007 and VOC2012 trainval set and testing on the VOC2007 test set. The experiment results show that the 2D detector achieves consistent performance gain via the combination of the proposed modules, which demonstrate the  versatility ability of our model.

\begin{table}[h]
 \begin{center}
 \begin{tabular}{|C{2.5cm}|C{2.2cm}|C{2.2cm}|}
  \hline
  Method & data & mAP\\
  \hline
  SSD300 & 07++12 & 77.2 \\
  +ALFR+DIT & 07++12 & \textbf{78.0} \\
  \hline
  \end{tabular}
 \end{center}
 \caption{The detection results on PASCAL VOC2007 test set via the combination of our model and SSD \cite{liu2016ssd}.}
  \label{2D_detection}
\end{table}
\section{Conclusion}
We have proposed a dynamic feature reflecting network (DFR-Net).
The proposed ALFR module separates the appearance perception and object localization decoding streams to exploit and reflect reciprocal information between sub-tasks in a self-mutual manner.
Our DIT module further scores the features of sub-tasks in a self-learning manner and accordingly realigns the multi-task training process.
Extensive experiments on the KITTI dataset demonstrate the effectiveness and the efficiency of our DFR-Net. It is worth mentioning that DFR-Net ranks $1^{st}$ in the highly competitive KITTI monocular 3D object detection track.
Besides, ablations on 2D detector SSD verify the generalization ability of the proposed modules.
Our method can also be amazingly plug-and-play on several cutting-edge frameworks at negligible cost. In future work, we will apply the proposed modules to more cutting-edge 3D detection approaches and other area to further verify the general ability of our model.

\clearpage

{\small
\bibliographystyle{ieee_fullname}
\bibliography{egbib}
}

\end{document}